\documentclass[runningheads]{llncs}
\usepackage[T1]{fontenc}
\usepackage{graphicx,verbatim}
\usepackage{booktabs}
\usepackage{multirow}
\usepackage{amsmath}
\usepackage{pifont}
\usepackage{amssymb}
\usepackage{graphicx} 
\usepackage{wrapfig}  
\usepackage{bbding}
\usepackage{authblk}
\usepackage{hyperref}
\begin{document}
\title{Frequency-domain Multi-modal Fusion for Language-guided Medical Image Segmentation}
%
%
%

\author{Bo Yu\inst{1}\textsuperscript{\dag} \and
Jianhua Yang\inst{2}\textsuperscript{\dag} \and
Zetao Du\inst{3} \and
Yan Huang\inst{2,4} \and
Chenglong Li\inst{5} \and
Liang Wang\inst{2,4}\textsuperscript{\Envelope}}
\authorrunning{Yu et al.}
%
\institute{School of Computer Science and Technology, Anhui University \and
NLPR, MAIS, Institute of Automation, Chinese Academy of Sciences \and
School of Information Science and Technology, ShanghaiTech University \and
School of Artificial Intelligence, University of Chinese Academy of Sciences \and
School of Artificial Intelligence, Anhui University \\
\email{wangliang@nlpr.ia.ac.cn}}




\maketitle              

\renewcommand{\thefootnote}{\dag}
\footnotetext{These authors contributed equally to this work.}
\renewcommand{\thefootnote}{\textsuperscript{}}
\footnotetext{The code is available at https://github.com/demoyu123/FMISeg.}

\begin{abstract}
Automatically segmenting infected areas in radiological images is essential for diagnosing pulmonary infectious diseases. Recent studies have demonstrated that the accuracy of the medical image segmentation can be improved by incorporating clinical text reports as semantic guidance. However, the complex morphological changes of lesions and the inherent semantic gap between vision-language modalities prevent existing methods from effectively enhancing the representation of visual features and eliminating semantically irrelevant information, ultimately resulting in suboptimal segmentation performance. To address these problems, we propose a Frequency-domain Multi-modal Interaction model (FMISeg) for language-guided medical image segmentation. FMISeg is a late fusion model that establishes interaction between linguistic features and frequency-domain visual features in the decoder. Specifically, to enhance the visual representation, our method introduces a Frequency-domain Feature Bidirectional Interaction (FFBI) module to effectively fuse frequency-domain features. Furthermore, a Language-guided Frequency-domain Feature Interaction (LFFI) module is incorporated within the decoder to suppress semantically irrelevant visual features under the guidance of linguistic information. Experiments on QaTa-COV19 and MosMedData+ demonstrated that our method outperforms the state-of-the-art methods qualitatively and quantitatively.


\end{abstract}
\section{Introduction}

The technology of medical image segmentation (MIS) is crucial for delineating pathological areas of pulmonary infectious diseases, such as COVID-19. It facilitates the precise identification of lesions and greatly aids in diagnosis, treatment planning, and disease monitoring. With the rapid developments of deep learning, numerous MIS methods based on CNN (e.g., U-Net \cite{bib6} and U-Net++ \cite{bib7}) and hybrid CNN-Transformer (e.g., TransUNet \cite{bib5} and SwinUnet \cite{bib9}) architectures have been proposed for the segmentation of pulmonary lesions from radiological images. These methods substantially assist physicians in identifying and evaluating pulmonary structures and pathological anomalies. Despite the remarkable progress of these methods, the intricate morphological characteristics of lesion regions (e.g., shape, size, and blurry boundaries) still pose critical challenges to effectively boosting the segmentation accuracy of pulmonary lesions.

In clinical practice, pulmonary imaging is typically accompanied by clinical text reports, which provide detailed descriptions of lesion regions, including their position, shape, size, number, and other relevant characteristics. Inspired by the significant performance improvement achieved by integrating textual information with visual information in MedCLIP \cite{bib0}, the task of language-guided medical image segmentation (LMIS) has attracted increasing attention \cite{bib12,bib14,bib13,bib27,bib28,bib29,bib30,bib31}. This task involves providing a medical image along with its corresponding text and predicting segmentation masks of pulmonary lesions. By leveraging the semantic guidance from the text, LMIS approaches significantly improve segmentation performance compared to uni-modal methods \cite{bib6,bib7,bib5,bib9}. To bridge the semantic gap between the medical image and the text in multi-modal frameworks, various fusion strategies have been explored, where linguistic and visual features are integrated either within visual encoder blocks (early fusion) \cite{bib12,bib14,bib29} or decoder blocks (late fusion) \cite{bib13,bib27,bib28,bib30,bib31}. Among these methods, both unidirectional interaction \cite{bib13,bib27,bib28} and bidirectional interaction \cite{bib30,bib31} between two modalities through attention mechanisms, as well as language-guided adapters in SAM \cite{bib10} have been investigated to improve semantic alignment.

However, two key issues still hinder the accurate localization and segmentation of target lesions specified by textual descriptions. Firstly, \textit{the visual representation of the medical image lacks sufficient discriminative ability.} The complex morphological changes of lesions indicate that the extraction of distinguished features is important for accurate segmentation, especially for the small or subtle lesion regions. Compared to spatial-domain features, frequency-domain features can enrich visual representations by providing complementary structural and textural information, which is beneficial for MIS \cite{bib32}. Nevertheless, the integration of frequency-domain features into LMIS task has not been explored to date. Secondly, \textit{semantically irrelevant visual information cannot be effectively suppressed.} Although existing LMIS methods \cite{bib13,bib27,bib28,bib30,bib31} leverage attention mechanisms to integrate visual and linguistic features, they often struggle to distinguish the lesion areas described by the text from complex anatomical backgrounds. For instance, the cross-modal attention in LanGuideSeg \cite{bib13} is insufficient to eliminate irrelevant visual information within the decoder, leading to suboptimal segmentation performance.

In this study, we propose a Frequency-domain Multimodal Interaction framework (FMISeg) to address the aforementioned issues in LMIS task. FMISeg follows a late fusion strategy to integrate textual information into high frequency (HF) and low frequency (LF) features within decoder blocks. Specifically, FMISeg first utilizes wavelet transform to generate HF and LF images from the raw image. Then, HF and LF images are fed into a dual-branch encoder to extract corresponding HF and LF features. Since HF features contain fine-grained textural details, while LF features encode high-level semantic contexts, their combination can enhance the representation of the raw image and contributes to more accurate lesion segmentation. To this end, our method introduces a Frequency-domain Feature Bidirectional Interaction (FFBI) module to fuse HF and LF features before feeding them into a dual-branch decoder. Furthermore, to effectively model the interaction between linguistic and visual features, we propose a Language and Frequency-domain Feature Interaction (LFFI) module within the decoder. This module first establishes bidirectional interaction between linguistic and visual features through a cross-attention mechanism. The LFFI module subsequently utilizes the generated filter weights to reweight the output visual features, thereby suppressing semantically irrelevant visual information. We conduct experiments on the QaTa-COV19 \cite{bib19} and MosMedData+\cite{bib17} datasets to validate the effectiveness and state-of-the-art performance of our method.

\begin{figure}[t]
    \centering
    \includegraphics[width=0.85\linewidth]{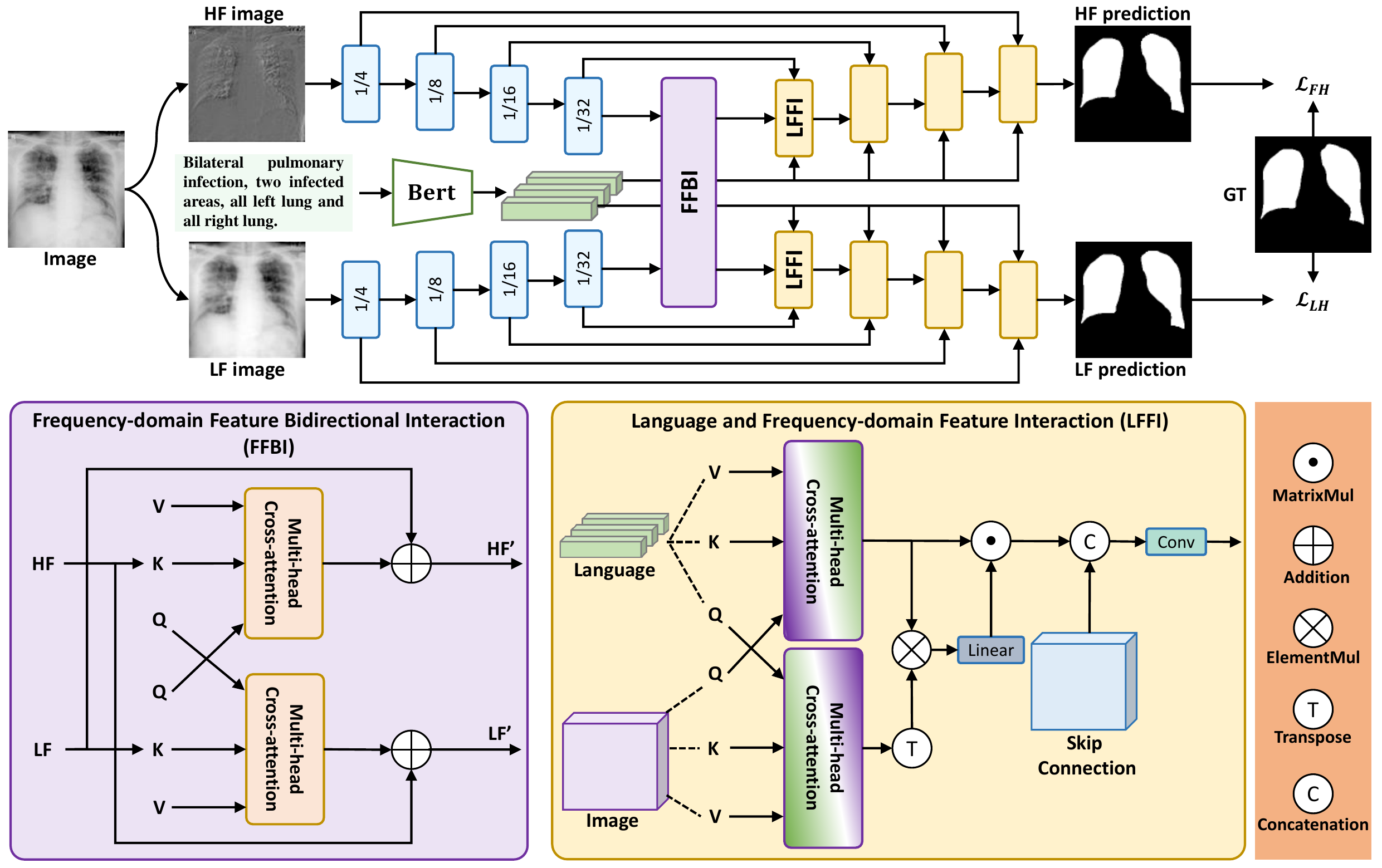}
    \caption{
Overview of the proposed frequency-domain multimodal interaction method.}
    \label{figure2}
\end{figure}

\section{Methodology}

An overview of our proposed method is illustrated in Fig. \ref{figure2}. The framework is comprised of four key components, including a dual-branch visual encoder to extract HF and LF features, a language encoder to extract linguistic features, the FFBI module to exchange supplementary information between HF and LF features at the final stage of the encoder, and a language-guided dual-branch decoder with LFFI modules to effectively align linguistic and visual modalities.

\subsection{Visual and Linguistic Feature Extraction}

\textbf{Vision Encoder:} For an input raw image $\mathbf{I} \in{\mathbb{R}^{H\times W\times3}}$, we follow XNet \cite{bib32} to apply wavelet transform to decompose it into the LF image $\mathbf{I}_{L}$ and the HF image $\mathbf{I}_{H}$. These LF and HF images are subsequently fed into dual-branch visual encoders (i.e., ConvNeXt-Tiny \cite{bib15}) to extract LF and HF visual features at different stages. Following previous works \cite{bib28,bib29,bib30,bib31}, we extract multi-stage LF and HF features with downsampling rates of 4, 8, 16, and 32. The extracted features are denoted as $\mathbf{F}_{m}^{1} \in \mathbb{R}^{\frac{H}{4} \times \frac{W}{4} \times C_{1}}$, $\mathbf{F}_{m}^{2} \in \mathbb{R}^{\frac{H}{8} \times \frac{W}{8} \times C_{2}}$, $\mathbf{F}_{m}^{3} \in \mathbb{R}^{\frac{H}{16} \times \frac{W}{16} \times C_{3}}$, and $\mathbf{F}_{m}^{4} \in \mathbb{R}^{\frac{H}{32} \times \frac{W}{32} \times C_{4}}$. Here,$H$ and $W$ represent the height and width of the input image, and $m \in \{LF, HF\}$, where $LF$ and $HF$ indicate low frequency and high frequency, respectively.          

\noindent \textbf{Language Encoder:} For the input clinical text $\mathbf{T} \in \mathbb{R}^{L}$, we follow prior works \cite{bib13,bib30,bib31} to adopt CXR-BERT \cite{bib16} as the language encoder to extract word-level linguistic features $\mathbf{F}_{T} \in \mathbb{R}^{L \times C}$. Here, $L$ and $C$ denote the number of words in the text and the channel dimension of features, respectively. Benefiting from its domain-specific optimization for chest X-ray reports, the linguistic features extracted from CXR-BERT can effectively facilitate semantic alignment between textual prompts and medical images.

\subsection{Frequency-domain Feature Bidirectional Interaction (FFBI)}

The LF features can encode high-level semantic contexts (e.g., organ morphology and lesion localization) and suppress noise interference. In contrast, HF features retain textural details (e.g., lesion and tissue boundaries) but are susceptible to HF artifacts and stochastic noise. To improve the segmentation of pulmonary lesions, we propose a Frequency-domain Feature Bidirectional Interaction (FFBI) module at the final stage of visual encoders, as illustrated in the bottom-left of Fig. \ref{figure2}. This module dynamically recalibrates HF features using LF semantic guidance, while simultaneously refines LF features with HF boundary details. Specifically, we leverage the cross-attention mechanism to model the bidirectional interactions between LF features $\mathbf{F}_{LF}^{4} \in \mathbb{R}^{\frac{H}{32} \times \frac{W}{32} \times C_{4}}$ and HF features $\mathbf{F}_{HF}^{4} \in \mathbb{R}^{\frac{H}{32} \times \frac{W}{32} \times C_{4}}$. This process can be formulated as:
\begin{equation}
    \hat{\mathbf{F}}_{HF}^{4} = LN(\mathbf{F}_{HF}^{4} + MHCA(\mathbf{F}_{HF}^{4}, \mathbf{F}_{LF}^{4}, \mathbf{F}_{LF}^{4})),
\end{equation}
\begin{equation}
    \hat{\mathbf{F}}_{LF}^{4} = LN(\mathbf{F}_{LF}^{4} + MHCA(\mathbf{F}_{LF}^{4}, \mathbf{F}_{HF}^{4}, \mathbf{F}_{HF}^{4})),
\end{equation}
where $LN(\cdot)$ denotes layer normalization, $MHCA(Q, K, V)$ represents multi-head cross-attention with inputs of query ($Q$), key ($K$), and value ($V$). Through bi-directional interaction, LF and HF features are respectively enhanced with local textural details and global semantic information. As a result, the proposed module yields more robust visual representations across diverse medical images.

\subsection{Language and Frequency-domain Feature Interaction (LFFI)}

To effectively model cross-modal interactions between linguistic features and LF (or HF) features, we propose a Language and Frequency-domain Feature Interaction (LFFI) module, as illustrated in the bottom-right of Fig. \ref{figure2}. Specifically, the LFFI module first establishes interactions between linguistic features and LF (or HF) features using two multi-head cross-attention structures. For clarity, we take LF features $\mathbf{F}_{LF}$ from arbitrary stages of the decoder as an example and omit the specific stage index. The process can be represented as:
\begin{equation}
   \mathbf{F}_{LF}^{'} = MHCA(\mathbf{F}_{LF}, \mathbf{F}_{T}, \mathbf{F}_{T}), 
\end{equation}
\begin{equation}
    \mathbf{F}_{T}^{'} = MHCA(\mathbf{F}_{T}, \mathbf{F}_{LF}, \mathbf{F}_{LF}).
\end{equation}

The interactions between linguistic and LF features via cross-attention are inadequate to bridge the semantic gap between modalities, as standard cross-attention may introduce semantically irrelevant noise and often struggles to capture fine-grained positional dependencies. To address this, we further design a semantically irrelevant filter to fuse $\mathbf{F}_{LF}^{'} \in \mathbb{R}^{h \times w \times C}$ and $\mathbf{F}_{T}^{'} \in \mathbb{R}^{L \times C}$, where $h$ and $w$ denote height and width of the feature map, respectively. Specifically, $\mathbf{F}_{LF}^{'}$ and $\mathbf{F}_{T}^{'}$ first perform the matrix multiplication and obtain the feature vector $\mathbf{F}_{M} \in \mathbb{R}^{h \times w \times L}$. Then, a linear projection layer followed by a sigmoid function is applied to generate the filter weights with the same dimensions as $\mathbf{F}_{LF}^{'}$. Finally, the weighted features resulting from the element-wise multiplication between the weights and $\mathbf{F}_{LF}^{'}$ are combined with encoder features via a skip connection to produce the output features $\mathbf{F}^{o}_{LF}$. This process can be described as:
\begin{equation}
    \mathbf{F}_{M} = \mathbf{F}_{LF}^{'} \otimes (\mathbf{F}_{T}^{'})^{\top},
\end{equation}
\begin{equation}
    \mathbf{F}_{LH}^{o} = Conv( \mathbf{F}_{LF} + \mathbf{F}^{'}_{LF} \odot \delta(Linear(\mathbf{F}_{M}))),
\end{equation}
where $\otimes$ denotes matrix multiplication, $\odot$ means element-wise multiplication, $\delta(\cdot)$ represents the sigmoid function, $Linear(\cdot)$ indicates linear projection, $Conv(\cdot)$ refers to convolution operation. Our dual-branch decoders progressively upsample the text-injected LF and HF features from high-level to low-level stages. Each branch independently predicts masks using two separate segmentation heads, which helps reduce learning bias. During training, we follow prior works \cite{bib12,bib31} to adopt Dice loss and cross-entropy loss to optimize dense predictions. 

\section{Experiments}
\subsection{Datasets and Evaluation Metrics}

To evaluate our method, we conducted experiments on two public medical image segmentation datasets with text prompts, including MosMeData+ \cite{bib17} and QaTa-COV19 \cite{bib19}. MosMedData+ contains 2,729 COVID-19 lung CT scan slices annotated with binary segmentation masks. It's divided into 2,183 training, 273 validation, and 273 test samples. \cite{bib12}. 
The QaTa-COV19 dataset contains 9,258 chest X-ray images, annotated with COVID-19 lesion details and textual descriptions of infection and location. It's split into 5,716 training, 1,429 validation, and 2,113 test samples. Following existing studies \cite{bib12,bib14}, we adopt two widely used evaluation metrics, namely the Dice coefficient ($Dice$) and mean intersection-over-union ($mIoU$), to evaluate the performance of the proposed method. 

\begin{table}[t!]
\centering
\small
\caption{{
Performance comparison of our FMISeg with existing medical image segmentation methods on the QaTa-COV19 and MosMedData+ datasets. 
}}
\setlength{\tabcolsep}{1.2mm}{
\begin{tabular}{l c c c c c c }
\toprule
\multirow{2}{*}{\textbf{Methods}} &\multirow{2}{*}{\textbf{Backbone}} & \multirow{2}{*}{\textbf{Text}}& \multicolumn{2}{c}{\textbf{QaTa-COV19}} & \multicolumn{2}{c}{\textbf{MosMedData+}} \\ \cmidrule(lr){4-7}
& & &Dice(\%)&mIoU(\%)&Dice(\%)&mIoU(\%)\\ 
\midrule
U-Net~\cite{bib6}       &CNN         & \ding{55} & 79.02 & 69.46 & 64.60 & 50.73 \\
UNet++~\cite{bib7}      &CNN         & \ding{55} & 79.62 & 70.25 & 71.75 & 58.39 \\
nnUnet~\cite{bib4}      &CNN         & \ding{55} & 80.42 & 70.81 & 72.59 & 60.36 \\
TransUNet~\cite{bib5}   &Hybrid      & \ding{55} & 78.63 & 69.13 & 71.24 & 58.44 \\
Swin-Unet~\cite{bib9}  &Hybrid      & \ding{55} & 78.07 & 68.34 & 63.29 & 50.19 \\
UCTransNet~\cite{bib23} &Hybrid      & \ding{55} & 79.15 & 69.60 & 65.90 & 52.69 \\
\midrule

LViT-T~\cite{bib12}        &Hydrid   & \ding{51} & 83.66 & 75.11 & 74.57 & 61.33 \\
LGA~\cite{bib14}           &Transformer & \ding{51} & 84.65 & 76.23 & 75.63 & 62.52 \\
CausalCLIPSeg~\cite{bib27} &Hybrid   & \ding{51} & 85.21 & 76.90 & - & - \\
RecLMIS~\cite{bib29}       &CNN      & \ding{51} & 85.22 & 77.00 & 77.48 & 65.07 \\
SGSeg~\cite{bib28}         &CNN      & \ding{51} & 87.41 & 77.82 & - & - \\
LanGuideSeg~\cite{bib13}   &CNN      & \ding{51} & 89.78 & 81.45 & - & - \\
MAdapter~\cite{bib30}      &CNN      & \ding{51} & 90.22 & 82.16 & 78.62 & 64.78 \\
TGCAM~\cite{bib31}         &CNN      & \ding{51} & 90.60 & 82.81 & 77.82 & 63.69 \\
\textbf{FMISeg (ours)} &CNN      & \ding{51} & \textbf{91.21} & \textbf{83.84} & \textbf{79.30} & \textbf{65.71} \\
\bottomrule
\end{tabular}
}
\label{result1}  

\end{table}

\subsection{Implementation Details}

We implemented our method in PyTorch \cite{bib20} with an NVIDIA RTX 3090 GPU. We adopted ConvNeXt-Tiny \cite{bib15} as the backbone for the dual-branch visual encoder. 
The model was optimized by the AdamW \cite{bib22} with an initial learning rate of 3e-4, which is eventually reduced to 1e-6, in conjunction with a cosine annealing learning rate strategy. For fair comparisons, the input image resolution was set to $224 \times 224$, and the default batch size was 32. The hidden dimension of the interaction module was set to 768.

\subsection{Comparison With State-of-the-Art Methods}

We compared the segmentation performance of our method with a series of state-of-the-art methods, including some classic uni-modal models (i.e., UNet \cite{bib6}, UNet++ \cite{bib7}, nnUNet \cite{bib4}, TransUNet \cite{bib5}, Swin-UNet \cite{bib9}, and UCTransNet \cite{bib23}) and all multi-modal models (i.e., LViT-T \cite{bib12}, LGA \cite{bib14}, CausalCLIPSeg \cite{bib27}, SGSeg \cite{bib28}, LanGuideSeg \cite{bib13}, RecLMIS \cite{bib29}, MAdapter \cite{bib30}, and TGCAM \cite{bib31}). The comparison results are shown in Table \ref{result1}. Note that the results of multi-modal methods are directly cited from their original papers, while the performance of uni-modal methods is taken from the reproduction provided by LViT \cite{bib12}. It can be observed that all methods without text show a significant performance gap compared to methods with text. For example, our method outperforms the best uni-modal nnUnet by 10.79\% in $Dice$ score and 13.03\% in $mIoU$ on the QaTa-COV19 dataset. When compared with LGA, which is based on foundation model SAM, our method has 6.56\% and 7.61\% improvement in terms of $Dice$ score and $mIoU$ on QaTa-COV19 dataset. Compared to methods using the same backbone, such as TGCAM, MAdapter, LanGuideSeg, and SGSeg, our method achieves superior performance on both datasets. Specifically, our method outperforms the best method (i.e., TGCAM) by 0.61\% in $Dice$ score and 1.03\% in $mIoU$ on QaTa-COV19 dataset, and by 1.48\% in $Dice$ score and 2.02\% in $mIoU$ on MosMedData+ dataset. 

\begin{figure}[t]
    \centering
    \includegraphics[width=0.90\linewidth]{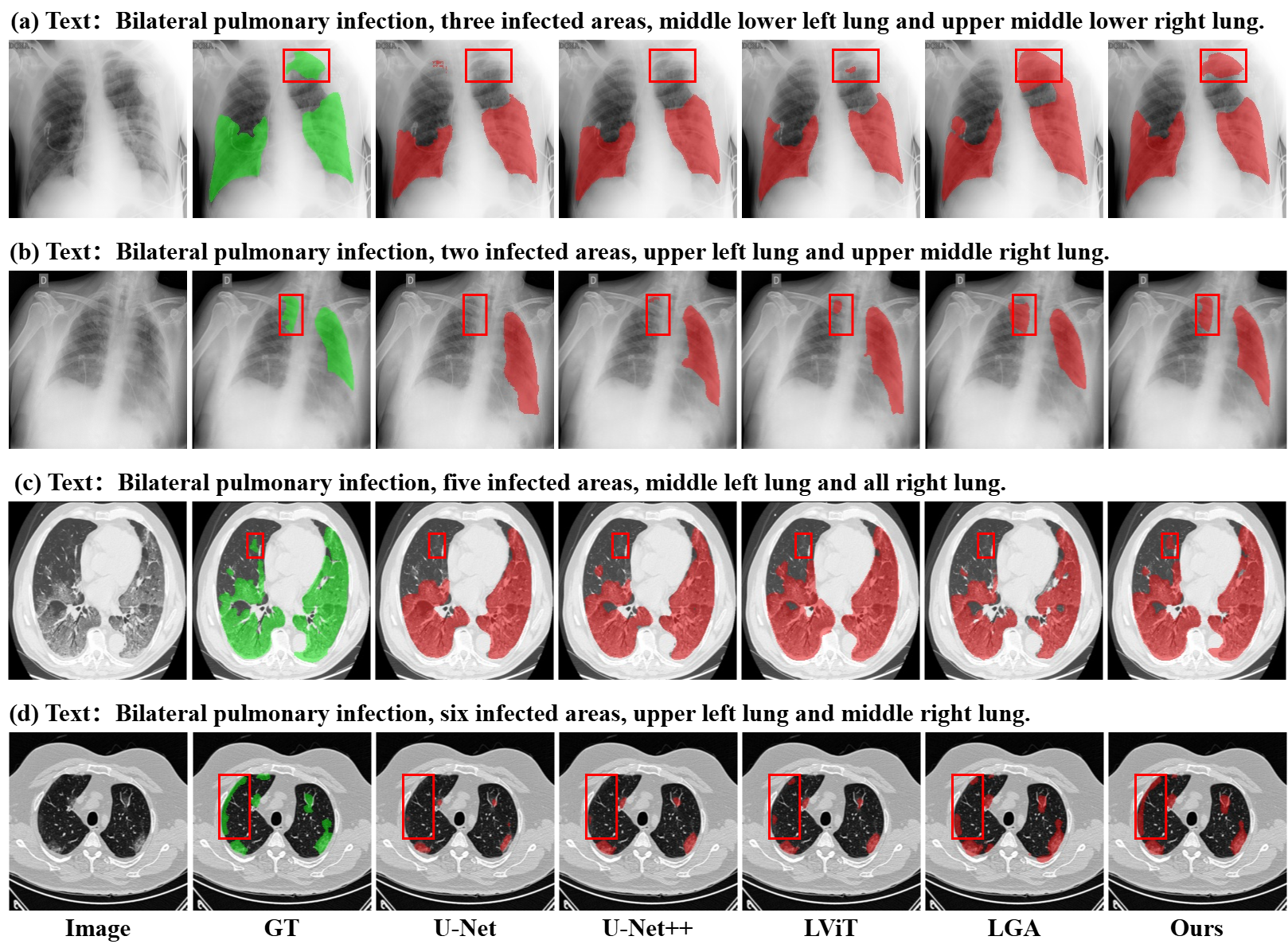}
    \caption{
Qualitative comparison of our method with uni-modal methods and multi-modal methods on QaTa-COV19 (a-b) and MosMedData+ (c-d).}
    \label{figure3}
\end{figure}
 
We also provide a qualitative comparison with uni-modal methods (i.e., U-Net and U-Net++) and multi-modal methods (i.e., LViT and LGA) in Fig. \ref{figure3}. It can be observed that uni-modal methods exhibit noticeable segmentation errors, particularly in boundary regions and detailed areas. While multi-modal methods show improved performance, they still suffer from missing or inaccurate segmentations in some complex and small areas. In contrast, our proposed method demonstrates higher accuracy, with more precise segmentation results.

\subsection{Ablation Study}
\begin{table}[h]
\centering
\caption{{
The effectiveness of the FFBI module.
}}
\setlength{\tabcolsep}{2.5mm}{
\begin{tabular}{l l c c c c c }
\toprule
\multirow{2}{*}{\textbf{No.}} &\multirow{2}{*}{\textbf{Model}} &\multicolumn{2}{c}{\textbf{QaTa-COV19}} & \multicolumn{2}{c}{\textbf{MosMedData+}} \\ 
\cmidrule(lr){3-6}
& &Dice(\%)&mIoU(\%)&Dice(\%)&mIoU(\%)\\ 
\midrule
\#1  & Raw Image    & 89.86 & 81.72 & 78.21 & 64.17 \\
\#2  & HF Image     & 88.75 & 80.15 & 77.16 & 63.04 \\
\#3  & LF Image     & 89.54 & 81.23 & 77.89 & 63.69 \\
\#4  & Cat(HF, LF)  & 90.61 & 82.93 & 78.65 & 64.88 \\
\#5  & FFBI(HF, LF) & \textbf{91.21} & \textbf{83.84} & \textbf{79.30} & \textbf{65.71} \\
\bottomrule
\end{tabular}
}
\label{result2}
\end{table}

\begin{table}[h]
\centering
\caption{{
The impact of the numbers of LFFI layers. The LFFI layers in each branch of the decoder range from 1 to 4. “No Text” means the absence of text in our method.
}}
\setlength{\tabcolsep}{2.5mm}{
\begin{tabular}{l l c c c c c }
\toprule
\multirow{2}{*}{\textbf{No.}} &\multirow{2}{*}{\textbf{Model}}& \multicolumn{2}{c}{\textbf{QaTa-COV19}} & \multicolumn{2}{c}{\textbf{MosMedData+}} \\ 
\cmidrule(lr){3-6}
& &Dice(\%)&mIoU(\%)&Dice(\%)&mIoU(\%)\\ 
\midrule
\#6  &No Text   & 87.63 & 78.13 & 76.45 & 62.87 \\
\#7  &1 layer   & 90.64 & 82.98 & 78.71 & 64.93 \\
\#8  &2 layers  & 90.86 & 83.26 & 78.97 & 65.25 \\
\#9  &3 layers  & 91.06 & 83.63 & 79.16 & 65.48 \\
\#10 &4 layers  & \textbf{91.21} & \textbf{83.84} & \textbf{79.30} & \textbf{65.71} \\
\bottomrule
\end{tabular}
}
\label{result3}
\end{table}

\noindent \textbf{Effectiveness of FFBI Module:} Our proposed method first decomposes the input raw image into LF and HF images, then establishes bidirectional interactions between LF and HF features in FFBI module. We conducted ablation studies to demonstrate the effectiveness of FFBI module, the experimental results are shown in Table \ref{result2}. The models of \#1, \#2, and \#3 are single-branch encoder-decoder models with inputs of raw, HF, and LF images, respectively. The models of \#4 and \#5 are two-branch encoder-decoder models, they adopt concatenation (Cat) and FFBI to fuse two modalities of visual features, respectively. The experimental results show that the performance of models fusing HF and LF visual features outperforms models using a single modality of visual features. This indicates that HF and LF modalities contain complementary information for segmentation. Furthermore, the FFBI module performs better than simply concatenating HF and LF features, since the bidirectional interaction with cross-attention is more conducive for mining complementary information from each other.

\noindent \textbf{The Impact of LFFI Layers:}
In our two-branch decoder, we progressively inject textual information using the LFFI module and upsample the fused visual features from high-level to low-level for mask prediction. Here, we conducted ablation studies to investigate the impact of LFFI layers on segmentation performance, experimental results are shown in Table \ref{result3}. These results show that Model \#6 performs the worst on two datasets when textual information is absent. However, it still significantly outperforms all uni-modal methods in Table \ref{result1}. As the layers of LFFI increase from high-level to low-level, the segmentation performance improves gradually. These experimental results demonstrate that the LFFI module effectively interacts with visual features and aligns the lesion region with textual semantics in our decoder.

\section{Conclusion}
In this paper, we propose a novel method FMISeg for language-guided medical image segmentation. FMISeg improves the segmentation accuracy of lesion regions from two aspects. Firstly, FMISeg extracts frequency-domain features and fuses different components of features with a frequency-domain feature bidirectional interaction module, resulting in the discriminative representation of visual features. Secondly, a language and frequency-domain feature interaction module is devised to effectively integrate linguistic features into frequency-domain features. Experiments show that FMISeg achieves state-of-the-art performance on the QaTa-COV19 and MosMedData+ datasets.

\hspace*{\fill}

\noindent \textbf{Acknowledgments.} This work was jointly supported by the National Natural Science Foundation of China (62236010, 62322607 and 62276261), and Youth Innovation Promotion Association of Chinese Academy of Sciences under Grant 2021128.

\hspace*{\fill}

\noindent \textbf{Disclosure of Interests.} The authors have no competing interests to declare that are relevant to the content of this paper.

%
%
%
%

\end{document}